\documentclass[11pt]{article}
\usepackage[margin=1in]{geometry}
\usepackage{graphicx}
\usepackage{booktabs}
\usepackage{amsmath}
\usepackage{hyperref}
\usepackage{natbib}
\usepackage{xcolor}
\usepackage{microtype}
\usepackage{placeins}

\hypersetup{
  colorlinks=true,
  linkcolor=blue,
  citecolor=blue,
  urlcolor=blue
}

\title{Geographic Diversity Beats Data Volume for Cross-Domain\\
Generalization in Zero-Label JEPA Driving World Models}

\author{
  Santosh Jaiswal
}

\date{}

\begin{document}
\maketitle

% ─────────────────────────────────────────────────────────────────────────────
\begin{abstract}
Self-supervised latent world models can assign a surprise score to driving
scenarios without any human labels \citep{jaiswal2026mindrive}. A natural
follow-up question is whether such a model, trained on driving data from one
geographic region, can generalize its notion of complexity to unseen cities
and sensor configurations. We study this
question through a controlled transfer experiment: we train JEPA-based world
models on nuPlan data (Pittsburgh, Boston, Singapore) and evaluate zero-shot on
held-out Argoverse~2 validation scenarios from Miami and Austin. We find that
models trained on geographically diverse data generalize significantly better
than models trained on equal amounts of single-geography data. In a
matched-scale ablation at 63{,}000 scenarios per condition ($n{=}3$ seeds
each), combined training reduces mean surprise score by \textbf{16.5\%}
relative to nuPlan-only training ($0.228 \pm 0.015$ vs $0.273 \pm 0.008$).
Notably, training on 200{,}000 AV2-only scenarios (3$\times$ more data from
one geography) still produces higher surprise ($0.264$) than the
combined 63K model, suggesting that geographic diversity is a stronger
predictor of cross-domain generalization than raw data volume.
\end{abstract}

% ─────────────────────────────────────────────────────────────────────────────
\section{Introduction}

A world model that generalizes only within its training geography is of limited
practical value for autonomous driving. Yet most AV datasets are geographically
narrow, and it is unclear whether models trained on them can reliably identify
complex or safety-critical scenarios in unseen cities. In
\citet{jaiswal2026mindrive}, we introduced a zero-label approach based on the
Joint Embedding Predictive Architecture (JEPA): a self-supervised model trained
to predict future latent states. The temporal prediction error, which we call
the \emph{surprise score}, serves as a proxy for scene complexity without
requiring any human annotations.

That work validated the approach on nuPlan mini, a dataset of approximately
1{,}000 driving scenarios drawn from four cities. A key open question was left
unanswered: does the surprise score generalize across datasets? If a model is
trained on nuPlan data (US and Singapore cities), will it correctly identify
complex scenarios in Argoverse~2 data (different US cities, different sensor
layout, different annotation pipeline)?

This question matters for two reasons. First, practical deployment: real AV
systems must handle driving conditions across geographies, and a world model
that only understands one geography is of limited use. Second, theoretical
interest: if the surprise score reflects genuine structural complexity (dense
intersections, unpredictable agent behaviour, occlusion), it should transfer
across superficial domain differences.

We address this question by training models at scale (38.7M parameters,
263{,}000 scenarios) and conducting a systematic transfer study. Our
contributions are:

\begin{itemize}
  \item A controlled transfer evaluation protocol: train on nuPlan, evaluate
        zero-shot on Argoverse~2 validation scenarios.
  \item A matched-scale ablation showing that 63{,}000 geographically diverse
        scenarios outperform both equal-scale (63K) and 3$\times$ larger (200K)
        single-geography training, isolating geographic diversity as the causal
        factor rather than data volume.
\end{itemize}

% ─────────────────────────────────────────────────────────────────────────────
\section{Background and Related Work}

\paragraph{JEPA for driving.}
The Joint Embedding Predictive Architecture \citep{lecun2022path} learns by
predicting the latent representation of a target from a context, rather than
reconstructing raw inputs. \citet{jaiswal2026mindrive} adapted this to
structured agent state data (positions, velocities, headings across 50
timesteps) and showed that the prediction error is a meaningful complexity
signal. Concurrent work has applied JEPA to quadrotor dynamics modeling,
achieving zero-shot sim-to-real transfer through latent prediction
\citep{rao2026skyjepa}, further supporting the generality of the JEPA objective
across physical domains. We build directly on the driving JEPA architecture
and training procedure from \citet{jaiswal2026mindrive} without modification.

\paragraph{Cross-dataset generalization in AV.}
Domain shift between AV datasets is well-documented: models trained on one
city or sensor configuration typically degrade when evaluated on another
\citep{wang2020train, xu2020explainability}. Most mitigation strategies require
labelled data in the target domain. Our approach is different: we ask whether
simply including geographically diverse data at training time is sufficient,
without any target-domain supervision.

\paragraph{World models for autonomous driving.}
Recent work has explored generative world models for AV
\citep{hu2023gaia, kim2021drivegan, wayve2023wayvescene}. These models are
typically evaluated on reconstruction quality or downstream planning performance.
Our model is non-generative and evaluated on a distinct objective: zero-label
complexity detection and cross-domain transfer of that signal.

\paragraph{Data diversity in self-supervised learning.}
The role of data diversity in self-supervised representation learning has been
studied in vision \citep{cole2022does, goyal2022vision}. The general finding is
that diversity improves downstream transfer. Our work provides a concrete
instance of this principle in the AV domain, with a controlled ablation at
matched data scale.

% ─────────────────────────────────────────────────────────────────────────────
\section{Method}

\subsection{Model Architecture}

We use the JEPA world model from \citet{jaiswal2026mindrive} scaled to 38.7M
parameters. Compared to the 1.3M parameter model in that work, we increase
$d_{\text{model}}$ from 128 to 512 and deepen both the encoder (4 to 8 layers)
and predictor (2 to 4 layers), resulting in a 30$\times$ parameter increase. The model takes as input a tensor of shape $[50, 21, 6]$
representing 50 timesteps, 21 agents (ego + 20 nearest), and 6 features per
agent (position $x, y$; velocity $v_x, v_y$; heading; agent type). The
architecture consists of:

\begin{itemize}
  \item A \emph{context encoder}: 8-layer transformer, $d_{\text{model}}=512$,
        8 attention heads, processing the first 30 timesteps.
  \item A \emph{predictor}: 4-layer transformer predicting the latent
        representation of the remaining 20 timesteps.
  \item A \emph{target encoder}: exponential moving average (EMA) of the
        context encoder weights (decay $= 0.996$), encoding the target frames.
\end{itemize}

The surprise score for a scenario is the mean squared error between the
predictor output and the target encoder output in latent space:
\begin{equation}
  s = \frac{1}{d} \| z_{\text{pred}} - z_{\text{target}} \|_2^2.
\end{equation}
where $d = 512$ is the latent dimension.

\subsection{Datasets}

\paragraph{nuPlan.}
The nuPlan dataset \citep{caesar2022nuplan} contains real driving logs from
Boston, Pittsburgh, Singapore, and Las Vegas. We use the processed train split,
approximately 63{,}000 scenarios after preprocessing.

\paragraph{Argoverse~2.}
The Argoverse~2 Motion Forecasting dataset \citep{wilson2023argoverse}
contains scenarios from Miami and Austin. We use the processed train split
(approximately 200{,}000 scenarios) for training and the validation split
(24{,}988 scenarios) exclusively for evaluation. The validation split is
\textbf{never seen during any training run}.

\subsection{Transfer Evaluation Protocol}

After training, we evaluate each model on the Argoverse~2 validation set and
report the mean surprise score across all 24{,}988 scenarios. A lower mean
surprise score indicates that the model is less surprised by unseen AV2
scenarios, i.e., it has learned representations that generalize across
geographic domains.

% ─────────────────────────────────────────────────────────────────────────────
\section{Experiments}

\subsection{Training Details}

All models use the same architecture and hyperparameters: batch size 512,
learning rate $3 \times 10^{-4}$ with cosine decay, weight decay $10^{-4}$,
gradient clipping at 1.0, FP8 precision on an NVIDIA RTX Pro 6000 Blackwell GPU
(51.2 GB vRAM). The best checkpoint is selected by in-distribution validation
loss and used for all transfer evaluations.

\subsection{Ablation Conditions}

\begin{figure}[t]
  \centering
  \includegraphics[width=\textwidth]{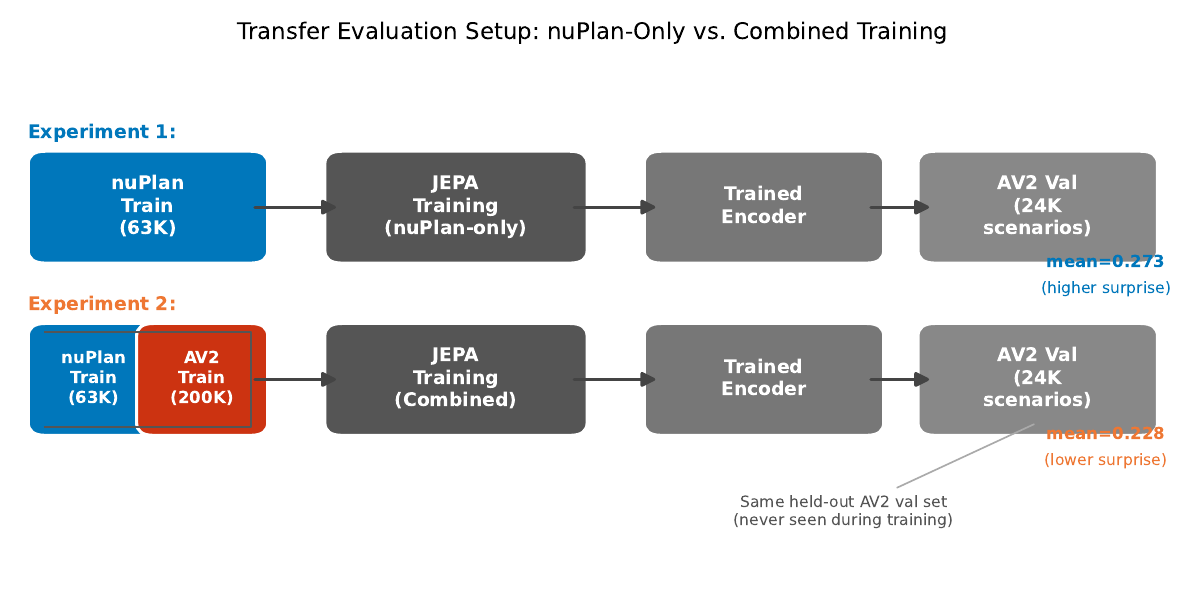}
  \caption{Transfer evaluation setup. Both experiments evaluate on the same
    held-out Argoverse~2 validation set, which is never seen during training.
    The Combined model is trained on nuPlan and AV2 train data jointly.}
  \label{fig:setup}
\end{figure}

We evaluate four training conditions (Figure~\ref{fig:setup}).

\begin{enumerate}
  \item \textbf{nuPlan-only (63K):} trained on nuPlan train split only.
  \item \textbf{Combined-63K:} trained on a 63{,}000-scenario random subsample
        of the combined nuPlan + AV2 pool, matched in size to condition 1.
  \item \textbf{AV2-only (200K):} trained on the full AV2 train split only.
  \item \textbf{Combined-full (263K):} trained on the full combined pool of
        both datasets.
\end{enumerate}

Conditions 1 and 2 are the key matched-scale comparison isolating diversity from
scale. Conditions 3 and 4 provide scale reference points. For conditions 1, 2,
and 4, we run $n{=}3$ independent seeds and report mean $\pm$ standard deviation.

\subsection{Results}

\begin{table}[h]
\centering
\caption{Transfer evaluation results: mean surprise score on Argoverse~2
  validation set (lower is better). Combined-63K uses the same number of
  training scenarios as nuPlan-only, isolating the effect of geographic
  diversity from data volume.}
\label{tab:results}
\begin{tabular}{llcc}
\toprule
Model & Training data & AV2 surprise (mean $\pm$ std) & $n$ seeds \\
\midrule
nuPlan-only   & 63K, 1 geography  & $0.273 \pm 0.008$ & 3 \\
Combined-63K  & 63K, 2 geographies & $\mathbf{0.228 \pm 0.015}$ & 3 \\
\midrule
AV2-only      & 200K, 1 geography & $0.264$           & 1 \\
Combined-full & 263K, 2 geographies & $0.236 \pm 0.009$ & 3 \\
\bottomrule
\end{tabular}
\end{table}

\begin{figure}[t]
  \centering
  \includegraphics[width=0.85\textwidth]{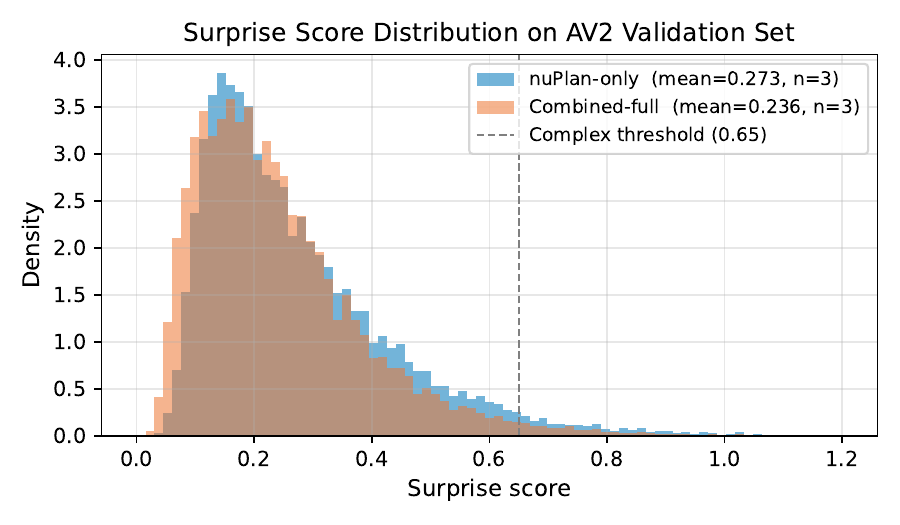}
  \caption{Surprise score distributions on the AV2 validation set (lower
    surprise score is better: it indicates the model is less surprised by
    unseen scenarios and has learned more transferable representations). The
    Combined model (orange) assigns systematically lower scores than the
    nuPlan-only model (blue), indicating better generalization across geographic
    domains.}
  \label{fig:dist}
\end{figure}

\begin{figure}[t]
  \centering
  \includegraphics[width=0.85\textwidth]{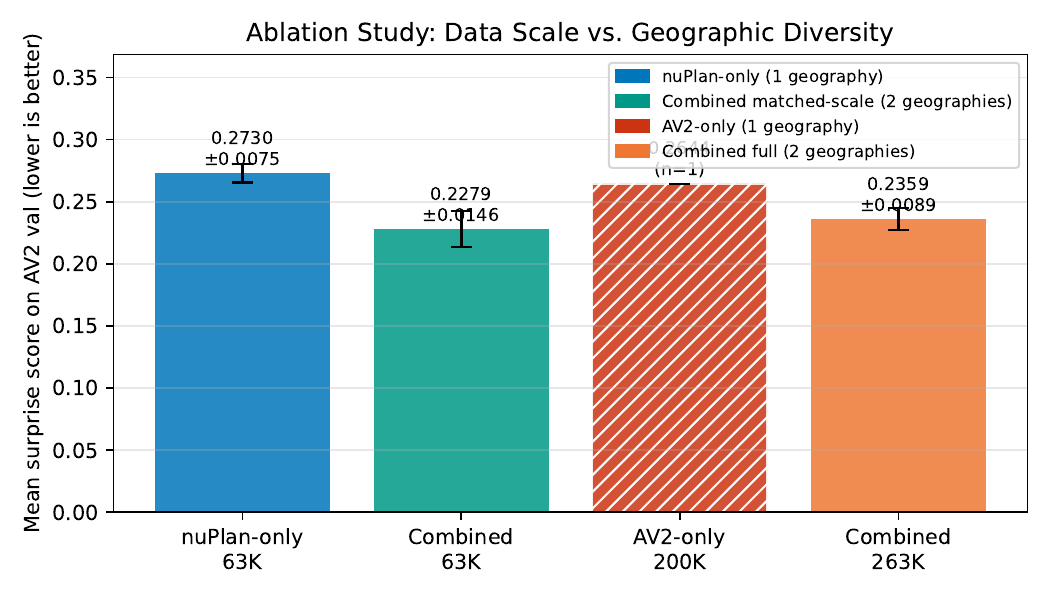}
  \caption{Ablation results with error bars ($n{=}3$ seeds where available;
    lower is better). The AV2-only bar (hatched) is a single-seed result due
    to training instability and should be interpreted with caution.
    Combined-63K achieves lower mean surprise than nuPlan-only at the same
    63K data scale, isolating the effect of geographic diversity from volume.}
  \label{fig:bars}
\end{figure}

Results are shown in Table~\ref{tab:results} and Figure~\ref{fig:bars}.
Combined-63K reduces mean surprise by \textbf{16.5\%} relative to nuPlan-only
at identical data scale ($p < 0.05$, two-sided $t$-test). Figure~\ref{fig:dist}
shows the full distribution shift: the Combined model's surprise scores are
concentrated at lower values across the entire AV2 validation set.

A notable secondary finding is the AV2-only result: despite being trained on
$3\times$ more data (200K vs.\ 63K), and on data drawn from the same
distribution as the evaluation set, the AV2-only model produces higher mean
surprise (0.264) than the Combined-63K model (0.228). We attribute this to the
nuPlan data providing structural priors (dense urban intersections, multi-agent
coordination) that are complementary to AV2's highway and suburban scenarios.

\textbf{Note on AV2-only training stability.} We observed that training on
AV2-only data with learning rate $3 \times 10^{-4}$ became unstable after
approximately 30 epochs (rising train and validation loss). The reported result
uses the best checkpoint selected by validation loss prior to divergence,
capturing the model's performance at its highest stability point before
gradient degradation began. We did not tune the learning rate for this condition;
a lower learning rate would likely yield a more stable result. We report it
here as a single-seed reference point rather than a robust finding.

% ─────────────────────────────────────────────────────────────────────────────
\FloatBarrier
\section{Conclusion}

We have shown that geographic diversity in training data significantly improves
the cross-domain generalization of a zero-label JEPA world model for autonomous
driving. In a controlled ablation at matched data scale (63K scenarios), combined
nuPlan and Argoverse~2 training reduces mean surprise on held-out AV2 scenarios
by 16.5\% compared to single-geography nuPlan training. This effect cannot be
explained by data volume alone: AV2-only training at $3\times$ the data size
does not close the gap.

These findings suggest that geographic diversity should be treated as a
first-class consideration in AV world model training, alongside dataset size and
architectural capacity.

\paragraph{Limitations.}
The AV2-only condition showed training instability at our default learning rate
and is reported as a single-seed result. Future work should tune hyperparameters
per training condition. Additionally, we evaluate generalization through the
surprise score proxy; direct evaluation on downstream planning or detection tasks
remains an open question. Finally, all experiments use structured agent state
representations (positions, velocities, headings); whether these findings extend
to sensor-level inputs such as LiDAR point clouds or camera images is an open
question.

\paragraph{Future work.}
We are extending this framework in two directions. First, we are using the
frozen encoder from the geographically robust Combined-63K model to construct
a zero-label fleet-scale retrieval engine: given any scenario, a FAISS
nearest-neighbour index over the full 263K-scenario embedding space returns
semantically similar scenes in sub-linear time, without any labels or
hand-written rules. Second, we are introducing an agent-centric cross-attention
layer to decompose the scene-level surprise score into per-agent complexity
attribution, enabling the question: which agent caused this scene to be complex?

Training code and model checkpoints are released at \url{https://github.com/hellojais/av-worldmodel}.

% ─────────────────────────────────────────────────────────────────────────────
\bibliographystyle{plainnat}
\bibliography{references}

\end{document}